\documentclass[sigconf]{acmart}

\usepackage{graphicx}
\usepackage{multirow}
\usepackage[switch]{lineno}
\AtBeginDocument{%
  \providecommand\BibTeX{{%
    \normalfont B\kern-0.5em{\scshape i\kern-0.25em b}\kern-0.8em\TeX}}}

\copyrightyear{2021}
\acmYear{2021}
\setcopyright{acmcopyright}\acmConference[ICMR '21]{Proceedings of the 2021 International Conference on Multimedia Retrieval}{August 21--24, 2021}{Taipei, Taiwan}
\acmBooktitle{Proceedings of the 2021 International Conference on Multimedia Retrieval (ICMR '21), August 21--24, 2021, Taipei, Taiwan}
\acmPrice{15.00}
\acmDOI{10.1145/3460426.3463641}
\acmISBN{978-1-4503-8463-6/21/08}

\acmSubmissionID{icmr164}

\begin{document}
\fancyhead{}
\title{Aligning Visual Prototypes with BERT Embeddings\\ for Few-Shot Learning}

\author{Kun Yan}
\email{kyan2018@pku.edu.cn}
\affiliation{%
  \institution{School of Software and Microelectronics, Peking University}
  \country{China}
}

\author{Zied Bouraoui}
\email{zied.bouraoui@cril.fr}
\affiliation{%
  \institution{CRIL - University of Artois  \& CNRS}
  \country{France}
}

\author{Ping Wang}
\authornote{means the corresponding author}
\email{pwang@pku.edu.cn}
\affiliation{%
  \institution{
   National Engineering Research Center for Software Engineering, Peking University}
  \country{China}
}
\additionalaffiliation{%
    \institution{the School of Software and Microelectronics, Peking University and the Key Laboratory of High Confidence Software Technologies (PKU), Ministry of Education}
}

\author{Shoaib Jameel}
\email{shoaib.jameel@gmail.com}
\affiliation{%
 \institution{School of Computer Science and Electronic Engineering, University of Essex}
 \country{UK}
}

\author{Steven Schockaert}
\email{schockaerts1@cardiff.ac.uk}
\affiliation{%
  \institution{School of Computer Science and Informatics, Cardiff University}
  \country{UK}
}
\renewcommand{\shortauthors}{Trovato and Tobin, et al.}

\begin{abstract}
  Few-shot learning (FSL) is the task of learning to recognize previously unseen categories of images from a small number of training examples. This is a challenging task, as the available examples may not be enough to unambiguously determine which visual features are most characteristic of the considered categories. To alleviate this issue, we propose a method that additionally takes into account the names of the image classes. While the use of class names has already been explored in previous work, our approach differs in two key aspects. First, while previous work has aimed to directly predict visual prototypes from word embeddings, we found that better results can be obtained by treating visual and text-based prototypes separately. Second, we propose a simple strategy for learning class name embeddings using the BERT language model, which we found to substantially outperform the GloVe vectors that were used in previous work. We furthermore propose a strategy for dealing with the high dimensionality of these vectors, inspired by models for aligning cross-lingual word embeddings. We provide experiments on miniImageNet, CUB and tieredImageNet, showing that our approach consistently improves the state-of-the-art in metric-based FSL.  
\end{abstract}





\maketitle

\section{Introduction}
Recent years have witnessed significant progress in image classification and related computer vision tasks~\cite{krizhevsky2012imagenet,simonyan2014very,goingdeeper,denseNet,resnext}, but most existing methods still require an abundance of labeled training examples. This stands in stark contrast with humans' ability to learn new categories from even a single example. This observation has fuelled research on designing systems that are capable of recognizing new image categories after only seeing a small number of examples, a task which is known as few-shot learning (FSL).  In this paper, we focus in particular on metric-based FSL methods~\cite{siamese,match-net,protonet,relationnet,few-shot-gnn}, which combine strong empirical performance with conceptual simplicity.

Metric-based methods aim to learn an embedding space which encourages generalization, i.e.\ where images from the same class are likely to have similar embeddings, even for unseen classes. An image can then be categorized based on its similarity to prototypes of the considered classes.
Despite significant progress in recent years, however, few-shot learning remains highly challenging. To alleviate the inherent difficulty of this task, some authors have proposed models that additionally take into account the name of the image classes. While these class names may not be available in all application settings, in those settings where they are, we can intuitively expect that they should provide us with meaningful prior knowledge.
Two notable examples of models that rely on class names are AM3 \cite{AM3} and TRAML \cite{traml}, both of which use the GloVe \cite{DBLP:conf/emnlp/PenningtonSM14} word embedding model for representing class names. %The underlying assumption, in both cases, is that categories with similar word vectors tend to be visually similar as well. In particular, 
In particular, the AM3 model tries to predict visual prototypes from the embeddings of the class names, while TRAML uses the similarity encoded by the word vectors to adapt the margin of the classifier.

However, standard word vectors, such as those from GloVe, are strongly influenced by topical similarity. This is illustrated in Table \ref{table7}, which shows the top-3 most similar classes from miniImageNet for three example targets. For instance, the nearest neighbours of \emph{catamaran} include \emph{snorkel} and \emph{jellyfish}. These words are all clearly topically related, but catamarans are not \emph{similar} to \emph{snorkels} or \emph{jellyfish}. This is problematic for few-shot learning, where we would intuitively want that class names with similar embeddings denote categories of the same kind. To address this issue, we propose a simple strategy for obtaining class name embeddings using the BERT masked language model \cite{DBLP:conf/naacl/DevlinCLT19}. We qualitatively observe that the resulting embeddings are indeed better suited for grouping classes that are conceptually similar. For instance, as can be seen in Table \ref{table7}, with the proposed BERT embeddings, the top 2 nearest neighbours are now also boats (being the only remaining boat classes in miniImageNet), while the third neighbour is also a vehicle. Furthermore, as the example of \emph{house finch} shows, the BERT embeddings also tend to model semantic relatedness at a finer-grained level: while the top neighbours for GloVe are all animals, none of them are birds. In contrast, the top two neighbours for BERT are birds. 

However, BERT embeddings also have the drawback of being  higher-dimensional: the BERT-large vectors on which we rely are 1024-dimensional, compared to 300 dimensions for the standard GloVe embeddings. This makes it difficult to predict visual prototypes from these vectors. Therefore, rather than predicting visual prototypes from the class names, we model the visual and text-based prototypes separately. Moreover, we also propose a dimensionality reduction strategy, inspired by work on aligning cross-lingual word embeddings \cite{DBLP:conf/aaai/ArtetxeLA18}, which aims to find a subspace of the BERT embeddings that is maximally aligned with the visual prototypes. As illustrated in Table \ref{table7}, the resulting embeddings remain at least as useful as the original BERT embeddings, despite only being 50-dimensional. In fact, some of the nearest neighbours for the low-dimensional vectors are arguably better than those of the BERT embeddings themselves, e.g.\ \emph{toucan} is more similar to \emph{house finch} than \emph{goose} is, while \emph{scoreboard} and \emph{street sign} are more meaningful neighbours of \emph{horizontal bar} than \emph{unicycle} and \emph{ear}.

\begin{table}[t]
\centering
\caption{Most similar miniImageNet classes to \emph{house finch}, \emph{horizontal bar} and \emph{catamaran}, according to class name embeddings obtained using GloVe, BERT and the proposed projection of the BERT embeddings onto a 50-dimensional space (BERT\textsubscript{proj}).}
%\footnotesize
%\resizebox{.95\columnwidth}{!}{
\begin{tabular}{lccc}
\toprule
%\noalign{\smallskip}
 & \textbf{catamaran} & \textbf{house finch} & \textbf{horizontal bar}  \\
%\noalign{\smallskip}
\midrule
%\noalign{\smallskip}
\multirow{3}{*}{GloVe} & snorkel & ladybug       & pencil box  \\
					   & yawl & komondor       & aircraft carrier\\
					    & jellyfish	& triceratops   & beer bottle \\
\midrule
\multirow{3}{*}{BERT}  & yawl & goose         & parallel bars\\
					    & aircraft carrier & toucan        & unicycle\\
						& school bus & ladybug       & ear \\
\midrule
\multirow{3}{*}{BERT\textsubscript{proj}} & yawl & toucan     & parallel bars \\
					      & school bus & robin      & scoreboard\\
						  & aircraft carrier & ladybug    & street sign\\ 
\bottomrule
\noalign{\smallskip}
\end{tabular}

\label{table7}
\end{table}

The main contributions of this paper are as follows: (i)
we propose a simple model for incorporating class names into metric-based FSL models, in which  visual prototypes and text-based prototypes are decoupled; (ii) we propose and evaluate several strategies for learning class name embeddings using BERT; (iii) we propose a strategy for dealing with the high dimensionality of the BERT embeddings by identifying the subspace of these embeddings which is most aligned with the visual prototypes.

\section{Related Work}
Most few-shot learning methods can be divided into metric-based  \cite{relationnet,edge_labeling,few-shot-gnn,feat}
and meta-learning based  \cite{optimization-as-model,MAML,Meta-SGD} methods, although some other directions have also been explored, such as hallucination based  \cite{low-shot-visual,low-shot-data,MetaGAN} and parameter-generation based  \cite{dynamic,implanting} methods.
Our focus in this paper is on metric-based methods, which essentially aim to learn a generalizable visual embedding space.
Early metric-based approaches used deep Siamese networks to compute the similarity between training and test images for the one-shot object recognition task \cite{siamese}. In these cases, a query image is simply assigned to the class of the most similar training image. Going beyond one-shot learning, \cite{match-net} proposed Matching Network, which uses a weighted nearest-neighbor classifier with an attention mechanism over the features of labeled examples.
Another important contribution of that work is the introduction of a new training scheme called episode-based learning, which uses a training procedure that is more closely aligned with the standard test setting for few-shot learning (see Section \ref{secProblemSetting}). The ProtoNet model from \cite{protonet} generates a visual prototype for each class, by simply averaging the embeddings of the available training images. The class of a query image is then predicted by computing its Euclidean distance to these prototypes. In the Relation Network \cite{relationnet}, rather than fixing the metric to be Euclidean, the model learns a deep distance metric to compare each query-support image pair.  In addition, some works have used Graph Convolutional Networks \cite{gcn} to exploit the relationship among support and query examples \cite{edge_labeling,few-shot-gnn}. The FEAT model, proposed by \cite{feat}, uses a transformer \cite{DBLP:conf/nips/VaswaniSPUJGKP17} to contextualize the image features relative to the support set in a given task. Recently, the Earth Mover's Distance~(EMD) has been adopted as a metric in DeepEMD~\cite{deepemd} to compute a structural distance between dense image representations to determine image relevance.
The aforementioned methods all rely on global image features. A few methods have also been proposed that aim to identify finer-grained local features, such as DN4~\cite{revisit}, SAML~\cite{saml}, STANet~\cite{dual-attention} and CTM~\cite{task-relevant}. 

The aforementioned methods only depend on visual features. A few methods also take into account the class names. In AM3 \cite{AM3}, prototypes are constructed as a weighted average of a visual prototype and a prediction from the class name. The relative weight of both modalities is computed adaptively and can differ from class to class. More recently, \cite{traml} used the class names as part of a margin based classification model. In this case, the underlying intuition is that a wider margin should be used for classes that have similar class names.
Within a wider scope, textual features have also been used for zero-shot image classification \cite{DBLP:conf/nips/FromeCSBDRM13,DBLP:conf/cvpr/ZhangXG17,DBLP:conf/cvpr/ChenZ00C18,Narayan2020}. %Existing approaches, however, either rely on standard word embedding models \cite{DBLP:conf/nips/FromeCSBDRM13,DBLP:conf/cvpr/ZhangXG17} or on explicit textual descriptions to estimate generative models of unseen classes \cite{DBLP:conf/cvpr/ChenZ00C18,Narayan2020}.
Recently, fuelled by the success of transformer based language models such as BERT \cite{DBLP:conf/naacl/DevlinCLT19}, a number of approaches have been proposed that train transformer models on joint image and text inputs, e.g.\ an image and its caption \cite{DBLP:conf/nips/LuBPL19,DBLP:conf/emnlp/TanB19,DBLP:conf/iclr/SuZCLLWD20}. Such models are aimed at tasks such as image captioning and visual question answering.

\section{Problem Setting}\label{secProblemSetting}

\begin{figure*}[t]
\centering
\includegraphics[width=0.95\textwidth]{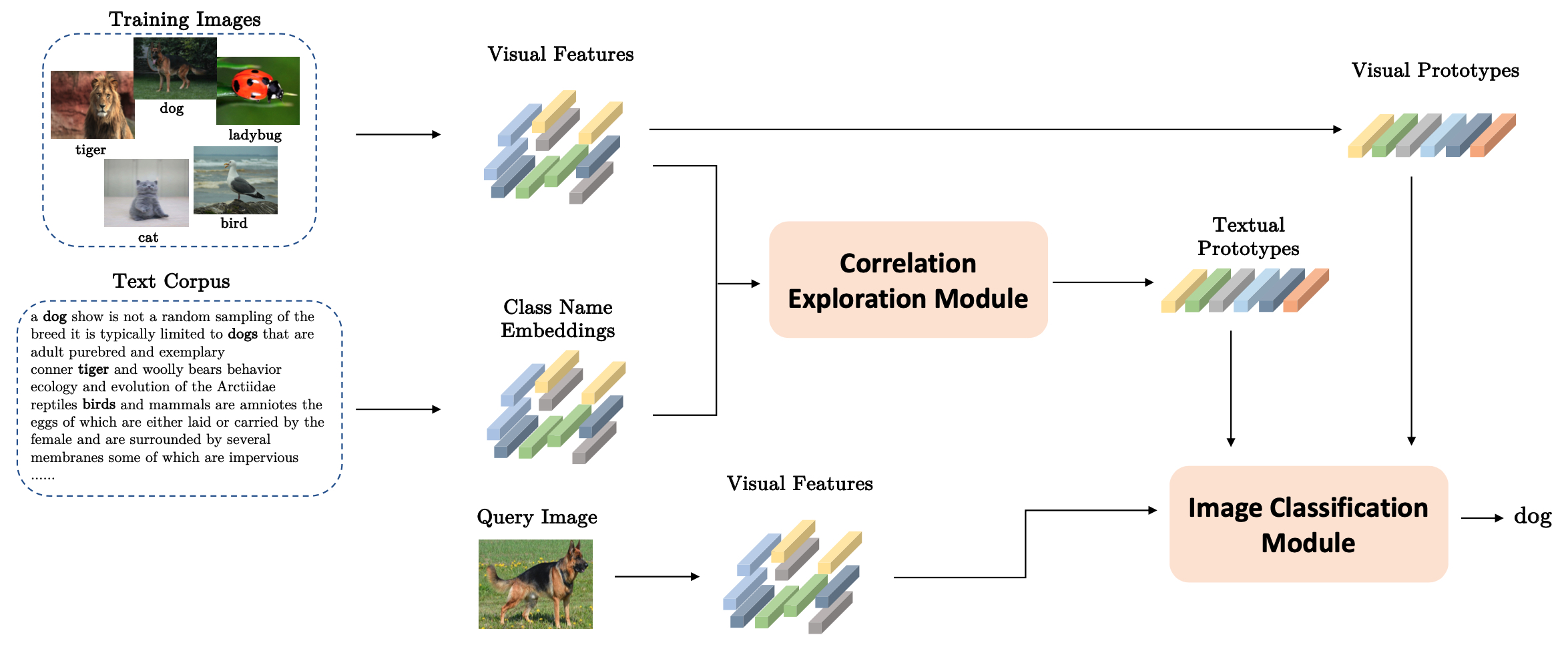} % Reduce the figure size so that it is slightly narrower  than the column.
\caption{Overview of our approach. After obtaining visual and text features, we use a Correlation Exploration Module to obtain visually meaningful low-dimensional textual prototypes. Both textual and visual prototypes are used in the final image classification step.}
\label{fig1}
\end{figure*}

In few-shot learning~(FSL), we are given a set of base classes $\mathcal{C}_{\textit{base}}$ and a set of novel classes $\mathcal{C}_{\textit{novel}}$, where $\mathcal{C}_{\textit{base}}~\cap~\mathcal{C}_{\textit{novel}} = \emptyset$. Each class in $\mathcal{C}_{\textit{base}}$ has sufficient labeled images, but for the classes in $\mathcal{C}_{\textit{novel}}$, only a few labeled examples are available. The goal of FSL is to obtain a classifier that performs well for the novel classes in $\mathcal{C}_{\textit{novel}}$. Specifically, in the $N$-way $K$-shot setting,  performance is evaluated using so-called episodes. In each test episode, $N$ classes from $\mathcal{C}_{\textit{novel}}$ are sampled, and $K$ labelled examples from each class are made available for training, where $K$ is typically 1 or 5. The remaining images from the sampled classes are then used as test examples. The support set of a given episode is the set of sampled training examples. We write it as $\mathcal{S} = \left\{ (x_{i}^{s}, y_{i}^{s}) \right\}_{i=1}^{n_{s}}$, where $n_{s} = N \times K$, $x_i^s$ are the sampled training examples and $y_i^s$ are the corresponding class labels. Similarly, the \emph{query set} contains the sampled test examples and is written as $\mathcal{Q} = \left\{ (x_{i}^{q}, y_{i}^{q}) \right\}_{i=1}^{n_{q}}$.

In this paper, we adopt the episode-based training scheme proposed by \cite{match-net}. In this case, the model is first trained by repeatedly sampling $N$-way $K$-shot episodes from $\mathcal{C}_{\textit{base}}$, rather than using $\mathcal{C}_{\textit{base}}$ directly. The way in which the training data from $\mathcal{C}_{\textit{base}}$ is presented thus resembles how the classifier is subsequently evaluated.

\section{Method}

The overview of our proposed architecture is shown in Fig.~\ref{fig1}. For a given episode, the labelled images are used to construct visual prototypes, as in existing approaches. Each of the class names is represented by a vector that was learned from some text corpus.
Both the visual prototypes and the class name embeddings feed into the Correlation Exploration Module~(CEM), whose aim is to find a low-dimensional subspace of the class name embeddings. The resulting textual prototype is then used in combination with the visual prototype for making the final prediction.

\subsection{Visual Features}
The visual features $f_{\theta}(x)\in \mathbb{R}^{n_{v}}$ of an image $x$ are extracted by a CNN model such as ResNet. 
Following ProtoNet~\cite{protonet}, in the $N$-way $K$-shot setting we construct the visual prototype of a class $c$ by averaging the visual features of all its training images in some episode $p$:
\begin{equation}
	\mathbf{v}^c_{p} = \frac{1}{K} \sum\{f_{\theta}(x_i^s) \,|\, (x_i^s,c)\in \mathcal{S}_p\}
\end{equation}
where $\mathcal{S}_p = \left\{ (x_{i}^{s}, y_{i}^{s}) \right\}_{i=1}^{n_{s}}$ is the support set of episode $p$.

\subsection{Class Name Embeddings}
We now explain how BERT \cite{DBLP:conf/naacl/DevlinCLT19} is used to get vector representations of class names. First note that BERT represents frequent words as a single token and encodes less common words as sequences of sub-word tokens, called word-pieces. Each of these tokens $t$ is associated with a static vector $\mathbf{t}\in \mathbb{R}^m$. 
The token vectors $\mathbf{t}$ are used to construct the initial representation of a given sentence $s=t_1,...,t_n$, which is subsequently fed to a deep transformer model. The output of this deep transformer model again consists of a sequence of token vectors, which intuitively represent the meaning of each token in the specific context of the given sentence. Let us write $m(s,i)$ for the output representation of $t_i$. When training BERT, some tokens of each input sentence are replaced by the special token [MASK]. If the token $t_i$ was masked, the output vector $m(s,i)$ acts as a prediction for the missing token.

Let $\mathcal{C}$ be the set of classes.  We first collect for each class $c\in \mathcal{C}$ a bag of sentences $S(c)={s_1,...,s_m}$ mentioning the name of this class. In particular, for each class name, we sample $m=1000$ such sentences from a given text corpus. We consider two strategies for learning class embeddings from these sentences. For the first strategy, we replace the entire class name by a single [MASK] token, and we use the corresponding output vector as the representation of $c$. We then take the average of the vectors we thus obtain across the $m$ sentences. In practice, the classes often correspond to WordNet synsets, meaning that we may have several synonymous names. In such cases, we first get a vector for each name from the synset (each learned from 1000 sentences), and then average the resulting vectors. The underlying assumption of this first approach is that when the i\textsuperscript{th} token is masked, the prediction $m(s,i)$ essentially encodes what the given sentence reveals about the meaning of the class $c$. This strategy has the important advantage that it can naturally deal with class names that consist of multiple word-piece tokens.
The second approach uses the full sentence as input, without masking any words. Following common practice \cite{pilehvar2019wic,he2020establishing}, the representation of $c$ is then obtained by averaging the output vectors of all the word-piece tokens corresponding to $c$.
We write $\mathbf{n}^c_{\textit{mask}}$ and $\mathbf{n}^c_{\textit{nomask}}$ for the embeddings obtained with the first and second method respectively. In addition to using $\mathbf{n}^c_{\textit{mask}}$ or $\mathbf{n}^c_{\textit{nomask}}$ individually, we will also experiment with their concatentation $\mathbf{n}^c_{\textit{mask}}\oplus \mathbf{n}^c_{\textit{nomask}}$. We will furthermore consider variants in which other types of word vectors are included, such as the GloVe embedding $\mathbf{n}^c_{\textit{glove}}$.

\subsection{Dimensionality Reduction}
One disadvantage of BERT embeddings is that they are high dimensional, a problem which is exacerbated when using concatenations of several types of class name embeddings. Furthermore, we can expect that only some of the information captured by the class name embeddings may be relevant for image classification. To address both shortcomings, we propose a Correlation Exploration Module (CEM), whose aim is to find a suitable lower-dimensional subspace of the class name embeddings.

Specifically, we aim to find linear mappings $\mathbf{A} \in \mathbb{R}^{m_t\times d}$ and $\mathbf{B} \in \mathbb{R}^{m_v\times d}$, where $m_t$ is the dimension of the class name embeddings, $m_v$ is the dimension of the visual features and $d < \min(m_t,m_v)$. Let $\mathbf{n}^c$ be the considered embedding of the name of class $c$, and let $\mathbf{v}^c_p$ be the visual prototype of the same class (for a given episode).
Intuitively, we want $\mathbf{n}^c \mathbf{A}$ to maximally retain the predictive information about $\mathbf{v}^c_p$ that is captured by $\mathbf{n}^c$. A natural strategy to find suitable matrices $\mathbf{A}$ and $\mathbf{B}$ is to use Canonical Correlation Analysis (CCA). These matrices are then chosen such that the correlations between the coordinates of $\mathbf{n}^c \mathbf{A} $ and the corresponding coordinates of $\mathbf{v}^c\mathbf{B}$ are maximized. The advantage of using CCA is that it is based on well-founded statistical principles and straightforward to compute. However, it was noted by \cite{DBLP:conf/aaai/ArtetxeLA18} that CCA is a sub-optimal choice for aligning cross-lingual word embeddings, which suggests that it may be a sub-optimal choice for cross-modal alignment as well. As pointed out in that paper, CCA can be seen as the combination of three linear transformations: (i) whitening of the initial vectors in the two embedding spaces,
%(i.e.\ transforming the vectors such that the individual components have unit variance and are uncorrelated), 
(ii) aligning the two spaces using orthogonal transformations and (iii) dimensionality reduction. It was found that better results can often be achieved by introducing an additional de-whitening step, which restores the original covariances. We will consider variants with and without this de-whitening step, which we will refer to as CCA+D and CCA respectively. The details of both variants are provided in the Appendix.

\subsection{Classification Model}
To classify a query image, we follow the set-up of ProtoNet, changing only the way in which the similarity between query images and prototypes is computed. In the case of ProtoNet, we have:
\begin{align}\label{eqProtoNet}
s_1(q,\mathbf{v}_p^c) = - \| f_{\theta}(q) - \mathbf{v}_p^c\|_2^2
\end{align}
The scores for each of the classes are then fed to a softmax layer to obtain class probabilities; the overall model is trained using the cross-entropy loss. In the case of FEAT, $f_{\theta}(q)$ and $\mathbf{v}_p^c$ are first contextualized using a transformer, before computing the squared Euclidean distance, as in \eqref{eqProtoNet}.

In our setting, we also have a class name embedding $\mathbf{n}^c$ for each class $c$. The most straightforward way of using this embedding is to estimate a mapping $g_{\psi}$ such that $g_{\psi}(\mathbf{n}^c)$ can be used as an approximation of the visual prototype $\mathbf{v}_p^c$. This is the strategy which is also pursued in AM3. However, instead of taking a weighted average of $\mathbf{v}_p^c$ and $g_{\psi}(\mathbf{n}^c)$ to obtain the final prototype, we keep the textual and visual prototypes separate. This allows us to use the cosine similarity to compare $g_{\psi}(\mathbf{n}^c)$ and $f_{\theta}(q)$, which has been found more suitable than Euclidean distance for comparing vectors that come from different distributions~\cite{dynamic}, while keeping the squared Euclidean distance for comparing $\mathbf{v}_p^c$ and $f_{\theta}(q)$. This leads to the following similarity score:
\begin{align}\label{eqDeepLearningModel}
s_2(q,\mathbf{v}_p^c) = - \| f_{\theta}(q) - \mathbf{v}_p^c\|^2_2 + \lambda \cos(f_{\theta}(q),g_{\psi}(\mathbf{n}^c))
\end{align}
where $\lambda$ is a hyper-parameter to control the contribution of the class name embeddings. To learn $g_{\psi}$, we use a shallow network consisting of a linear transformation onto a 512-dimensional layer with ReLU activation and batch normalization \cite{batchnormalization}, followed by another linear transformation.

As mentioned above, learning a suitable mapping $g_{\psi}$ is challenging when $\mathbf{n}^c$ is high-dimensional. Rather than learning the parameters of this mapping as part of the model, we therefore propose to use the mappings $\mathbf{A}$ and $\mathbf{B}$ that were found by the Correlation Exploration Module. The similarity score thus becomes:
\begin{align*}
s_3(q,\mathbf{v}_p^c) = - \| f_{\theta}(q) - \mathbf{v}_p^c\|^2_2 + \lambda \cos(f_{\theta}(q)\mathbf{B},\mathbf{n}^{c}\mathbf{A})
\end{align*}

\section{Experiments}

\subsection{Experimental Setup}
\subsubsection{Datasets} We conduct experiments on three benchmark datasets: miniImageNet~\cite{match-net}, tieredImageNet~\cite{tiered} and CUB~\cite{CUB}. MiniImageNet is a subset of the ImageNet dataset~\cite{ImageNet}. It consists of 100 classes, each with 600 labeled images of size 84 $\times$ 84. We adopt the common setup introduced by~\cite{optimization-as-model}, which defines a split of 64, 16 and 20 classes for training, validation and testing respectively. TieredImageNet is a larger-scale dataset with more classes, containing 351, 97 and 160 classes for training, validation and testing. The CUB dataset contains 200 classes and 11\,788 images in total. We used the splits from \cite{closer-look}, where 100 classes are used for training, 50 for validation, and 50 for testing.

\subsubsection{Training and Test Setting} 
We evaluate our method on 5-way 1-shot and 5-way 5-shot settings. We train 50\,000 episodes in total for miniImageNet, 80\,000 episodes for tieredImageNet and 40\,000 episodes for CUB. During the test phase, 600 test episodes are generated. We report the average accuracy as well as the corresponding 95\% confidence interval over these 600 episodes.

\subsubsection{Class Name Embeddings}
As baseline class name embedding strategies, we used 300-dimensional FastText  \footnote{\url{https://fasttext.cc/docs/en/crawl-vectors.html}}\cite{bojanowski2017enriching}, GloVe \footnote{\url{https://nlp.stanford.edu/projects/glove/}} \cite{DBLP:conf/emnlp/PenningtonSM14} and skip-gram embeddings\footnote{\url{https://code.google.com/archive/p/word2vec/}} \cite{mikolov2013distributed}. For the BERT embeddings, we use the BERT-large-uncased model\footnote{Available from \url{https://github.com/huggingface/transformers}}, which yields 1024 dimensional vectors.
To obtain the $\mathbf{n}^c_{\textit{mask}}$ and $\mathbf{n}^c_{\textit{nomask}}$ vectors, we used the May 2016 dump of the English Wikipedia.  In addition to using the vectors $\mathbf{n}^c_{\textit{mask}}$ (referred to as BERT\textsubscript{mask}) and $\mathbf{n}^c_{\textit{nomask}}$ (referred to as BERT\textsubscript{nomask}), we also experiment with the following concatenations: $\mathbf{n}^c_{\textit{mask}}\oplus\mathbf{n}^c_{\textit{nomask}}$ (referred to as CON$_1$) and  $\mathbf{n}^c_{\textit{mask}}\oplus\mathbf{n}^c_{\textit{nomask}}\oplus \mathbf{n}^c_{\textit{glove}}$ (referred to as CON$_2$).

\begin{table}[t]
\centering
\caption{Comparison of the performance of different word embedding models on miniImageNet, for the 5-way 5-shot setting using the learned mapping network $g_{\psi}$.}
%\footnotesize
%\resizebox{.95\columnwidth}{!}{
\begin{tabular}{lcc}
\toprule
\textbf{Word Emb.} & $\mathbf{t_{\textbf{max}}}$ & \textbf{Accuracy} \\

\midrule
FastText       &   & 74.97 +- 0.65  \\
GloVe          &   & 75.30 +- 0.61 \\
Skip-gram      &  & 74.91 +- 0.66  \\
\midrule
BERT\textsubscript{static} & & 74.53 +- 0.67\\
\midrule
BERT\textsubscript{mask} & 16     & 75.47 +- 0.68  \\
BERT\textsubscript{mask} & 32     & 75.86 +- 0.61  \\
BERT\textsubscript{mask} & 64    & {\bf 76.30 +- 0.76}  \\
BERT\textsubscript{mask} & 100   & 75.50 +- 0.63  \\
\midrule
BERT\textsubscript{nomask} & 16  & 74.73 +- 0.66  \\
BERT\textsubscript{nomask} & 32  & 74.79 +- 0.67  \\
BERT\textsubscript{nomask} & 64  & 75.62 +- 0.65  \\
BERT\textsubscript{nomask} & 100 & 74.76 +- 0.69  \\
\bottomrule
\noalign{\smallskip}
\end{tabular}
\label{table1}
\end{table}

\subsubsection{Implementation Details}
We have implemented\footnote{\url{https://github.com/yankun-pku/Aligning-Visual-Prototypes-with-BERT-Embeddings-for-Few-shot-Learning}} our model using the PyTorch-based framework provided by \cite{closer-look}. As the backbone network for  the visual feature embeddings, we used \emph{ResNet-10}~\cite{residual} for the ablation study in Section \ref{secAblationStudy} and \emph{ResNet-12} 
%, \emph{ResNet-18} 
and Conv-64~\cite{protonet} for our comparison with the state-of-the-art in Section \ref{secComparison}. Conv-64 is the standard choice for CUB. It has four layers with each layer consisting of a 3 $\times$ 3 convolution and filters, followed by batch normalization, a ReLU non-linearity, and 2 $\times$ 2 max-pooling. All experiments are trained from scratch using the Adam optimizer with an initial learning rate of 0.001. In experiments where the mapping network $g_{\psi}$ is used, this network is trained separately, with a learning rate of 0.0001. The remaining parameters are selected based on the validation set. In particular, the coefficient $\lambda$ is chosen from $\{1,2,...,10\}$. For miniImageNet and CUB, the optimal value was $\lambda=5$; for tieredImageNet we obtained $\lambda=6$. We similarly select the type of class name embedding from 
% $\{$BERT\textsubscript{mask}, CON$_1$, CON$_2$, CON$_3$, CON$_4\}$ 
$\{$BERT\textsubscript{mask}, CON$_1$, CON$_2\}$ 
and the number of dimensions from $\{25,50,100,200\}$. In all cases, we used the CCA+D method for reducing the number of dimensions. For miniImageNet, 50-dimensional CON$_2$ was selected; for CUB, 50-dimensional CON$_1$ was selected; for tieredImageNet, 100-dimensional CON$_2$ was selected.

\subsection{Ablation Study}\label{secAblationStudy}
Our ablation study is based on the ProtoNet model. 
All experiments in this section are conducted on miniImageNet using ResNet-10 as the feature extractor.

\subsubsection{Word Embedding Models} 
%\subsubsection{Word Embedding Models}
We first explore the impact of the considered word embedding model. We found that the BERT-based approach is sensitive to sentence segmentation errors. To mitigate the impact of such errors, we only considered sentences whose length is below a maximum of $t_{\textit{max}}$ word-piece tokens, where we considered values of $t_{\textit{max}}$ between 16 and 100. The results are shown in Table \ref{table1}, where we used the variant of our model with the learned mapping network $g_{\psi}$ for 5-way 5-shot learning. The results show that BERT\textsubscript{mask} consistently outperforms BERT\textsubscript{unmask}, while $t_{\textit{max}}=64$ achieves the best balance between avoiding sentences with segmentation issues and removing too many sentences. BERT\textsubscript{mask} performs consistently better than GloVe, which achieves the best performance among the baseline models. The static BERT input vectors (shown as BERT\textsubscript{static}) achieve the worst performance overall. In the remainder of the experiments, we fix $t_{\textit{max}}=64$.

\begin{table}[t]
\centering
%\footnotesize
\caption{Results for textual prototypes of different dimensionality on miniImageNet, for the 5-way 5-shot setting.}
%\resizebox{.95\columnwidth}{!}{
\begin{tabular}{ccc}
\toprule
\textbf{Dim} & \textbf{CCA} & \textbf{CCA+D}\\

\midrule
25            & 76.21 +- 0.62 &75.99 +- 0.64  \\
50            & 76.17 +- 0.67 &{\bf 76.40 +- 0.63} \\
100           & 75.91 +- 0.66 & 76.32 +- 0.65  \\
200  & 75.98 +- 0.68 & 76.17 +- 0.64\\
\bottomrule
\noalign{\smallskip}
\end{tabular}

\label{table2}
\end{table}

\subsubsection{Correlation Exploration Module}
%\subsubsection{Correlation Exploration Module}
We now analyze the usefulness of the Correlation Exploration Module, comparing in particular the CCA and CCA+D alignment strategies. Note that when the mapping network $g_{\psi}$ is used we are forced to keep the dimensionality the same as that of the visual features (which is 512 in the case of ResNet), whereas with the CCA-based alignment methods, we can use lower-dimensinal textual prototypes. Table \ref{table2} explores the effect of the dimensionality $d$ of the textual prototypes. The best results were found for $d=50$. The results for $d=50$ are similar to the results we obtained with the mapping network $g_{\psi}$ in Table \ref{table1}, with CCA+D performing slightly better and CCA performing slightly worse than BERT\textsubscript{mask}.

\begin{table}[t]
%\footnotesize
\centering
\caption{Comparison of different alignment strategies on miniImageNet, for the 5-way 5-shot setting, with $d=50$.}
%\resizebox{.95\columnwidth}{!}{
\begin{tabular}{lll}
\toprule
\textbf{Alignment Method} & \textbf{Word Emb.}  & \textbf{Accuracy} \\

\midrule
$g_{\psi}$     & BERT\textsubscript{mask} & 76.30 +- 0.76\\
$g_{\psi}$     & CON$_1$   &  75.72 +- 0.60    \\
$g_{\psi}$     & CON$_2$   &  75.16 +- 0.79    \\
% $g_{\psi}$     & CON$_3$   &  75.44 +- 0.63    \\
% $g_{\psi}$    & CON$_4$   &  75.23 +- 0.62 \\
 \midrule
 CCA     & BERT\textsubscript{mask}  &  76.21 +- 0.62   \\
 CCA     & CON$_1$             &  76.31 +- 0.67  \\
 CCA     & CON$_2$             &  76.50 +- 0.62  \\
%  CCA     & CON$_3$             &  76.23 +- 0.65  \\
% CCA     & CON$_4$     &  76.49 +- 0.66  \\
\midrule
CCA+D     & BERT\textsubscript{mask}   &  76.40 +- 0.63   \\
CCA+D     & CON$_1$             &  76.61 +- 0.65\\
CCA+D     & CON$_2$             &   {\bf 76.82 +- 0.64}\\
% CCA+D     & CON$_3$             &   76.59 +- 0.61\\
% CCA+D     & CON$_4$    &  76.60 +- 0.62 \\
\bottomrule
\noalign{\smallskip}
\end{tabular}
\label{table3}
\end{table}

\begin{figure}[t]
\centering
\includegraphics[width=0.9\columnwidth]{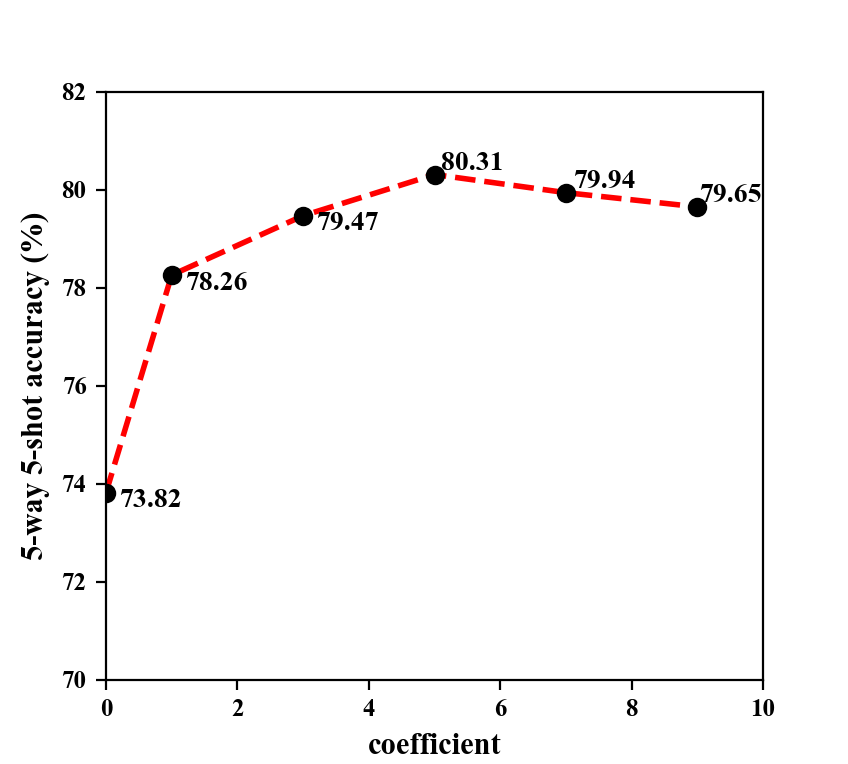} % Reduce the figure size so that it is slightly narrower than the column. Don't use precise values for figure width.This setup will avoid overfull boxes.
\caption{5-way 5-shot accuracy with different $\lambda$ values on the miniImageNet validation dataset.}
\label{fig2}
\end{figure}

However, a key advantage of the CCA methods is that we can further increase the dimensionality of the class name embeddings, without increasing the number parameters of the classification model. To further explore the potential of these alignment methods, Table \ref{table3} shows the results for different concatenations, each time keeping the dimensionality of the textual prototypes fixed at $d=50$. As can be seen, when the mapping network $g_{\psi}$ is used, these concatenations degrade the performance, as the high dimensionality of the input vectors leads to overfitting. In contrast, with CCA and CCA+D we see some clear performance gains, where CCA+D again outperforms CCA. Among the different concatenation strategies, CON$_2$ performs best.

\subsubsection{Coefficient $\lambda$}
The hyper-parameter $\lambda$ controls the contribution of the textual prototypes to the overall similarity computation. Figure \ref{fig2} shows the impact of this coefficient on the accuracy of the validation set from miniImageNet, where the BERT\textsubscript{mask} vectors with the CCA+D alignment strategy were used. In this case, the best results are found for $\lambda=5$. Note that $\lambda=0$ corresponds to the standard ProtoNet model, which achieves the worst results within the considered range of $\lambda$.

\begin{table}[t]
%\footnotesize
\begin{center}
\caption{Comparison with AM3 on miniImageNet (using ResNet-12 in all cases), showing mean accuracies (\%) with a 95\% confidence interval.}
\begin{tabular}{llcc}
\multicolumn{4}{l}{\textbf{5-way 1-shot setting:}} \\
\toprule
\textbf{Word Emb.} & \textbf{Base Met.} & \textbf{AM3} & \textbf{Ours}\\
\midrule
GloVe     & ProtoNet  & 62.43 $\pm$ 0.80 & 63.49 $\pm$ 0.67\\
BERT      & ProtoNet  & 62.11 $\pm$ 0.39 & 63.84 $\pm$ 0.32\\
CON$_1$   & ProtoNet  & 62.14 $\pm$ 0.41 & 64.13 $\pm$ 0.45\\
CON$_2$   & ProtoNet  & 62.03 $\pm$ 0.46 & 64.53 $\pm$ 0.37\\
\bottomrule

\noalign{\smallskip}
\noalign{\smallskip}

\multicolumn{4}{l}{\textbf{5-way 5-shot setting:}} \\
\toprule
%\noalign{\smallskip}
\textbf{Word Emb.} & \textbf{Base Met.} & \textbf{AM3} & \textbf{Ours}\\
%\noalign{\smallskip}
\midrule
%\noalign{\smallskip}
GloVe     & ProtoNet  & 74.87 $\pm$ 0.65 & 78.72 $\pm$ 0.64\\
BERT\textsubscript{mask}      & ProtoNet  & 74.72 $\pm$ 0.64 & 79.10 $\pm$ 0.63\\
CON$_1$   & ProtoNet  & 74.24 $\pm$ 0.68 & 79.26 $\pm$ 0.65\\
CON$_2$   & ProtoNet  & 74.09 $\pm$ 0.70 & 79.37 $\pm$ 0.64\\
\bottomrule

\noalign{\smallskip}
% \midrule
\end{tabular}

\label{table8a}
\end{center}
\end{table}

\begin{table}[t]
%\footnotesize
\begin{center}
\caption{Comparison with TRAML on miniImageNet (using ResNet-12 in all cases), showing mean accuracies (\%) with a 95\% confidence interval.}
\begin{tabular}{llcc}
\multicolumn{4}{l}{\textbf{5-way 1-shot setting:}} \\
\toprule
\textbf{Word Emb.} & \textbf{Base Met.} & \textbf{TRAML} & \textbf{Ours}\\
\midrule
%\noalign{\smallskip}
GloVe     & ProtoNet         & 60.31 $\pm$ 0.48 & 63.49 $\pm$ 0.67\\
GloVe     & AM3(ProtoNet)    & 67.10 $\pm$ 0.52 & 67.75 $\pm$ 0.39\\
CON$_2$   &  AM3(ProtoNet)   & -                & 68.42 $\pm$ 0.51\\
\bottomrule

\noalign{\smallskip}
\noalign{\smallskip}
\multicolumn{4}{l}{\textbf{5-way 5-shot setting:}} \\
\toprule
\textbf{Word Emb.} & \textbf{Base Met.} & \textbf{TRAML} & \textbf{Ours}\\
\midrule
%\noalign{\smallskip}
GloVe     & ProtoNet         & 77.94 $\pm$ 0.57 & 78.72 $\pm$ 0.64\\
GloVe     & AM3(ProtoNet)    & 79.54 $\pm$ 0.60 & 80.62 $\pm$ 0.76\\
CON$_2$   & AM3(ProtoNet)    & -                & 81.29 $\pm$ 0.59\\
\bottomrule
\noalign{\smallskip}
% \midrule
\end{tabular}
\label{table8b}
\end{center}
\end{table}

\begin{table*}[t]
%\footnotesize
\begin{center}
\caption{The mean accuracies~(\%) with a 95\% confidence interval %of the 5-way 5-shot and 5-way 1-shot tasks 
on the miniImageNet dataset.}
\begin{tabular}{lllcc}
\toprule
%\noalign{\smallskip}
\textbf{Method} & \textbf{Backbone} & \textbf{Type}
&\textbf{5-way 1-shot} &\textbf{5-way 5-shot}\\
%& & & 1-shot & 5-shot
%\noalign{\smallskip}
\midrule
%\noalign{\smallskip}
MAML~\cite{MAML}   	        & Conv-64 & Meta     & 48.70 $\pm$ 1.75 & 63.15 $\pm$ 0.91\\
Reptile~\cite{Reptile}  	    & Conv-64 & Meta     & 47.07 $\pm$ 0.26 & 62.74 $\pm$ 0.37\\
LEO~\cite{meta-latent-embedding} & WRN-28 & Meta   & 61.76 $\pm$ 0.08 & 77.59 $\pm$ 0.12\\
MTL~\cite{mtl}  	& ResNet-12 & Meta     & 61.20 $\pm$ 1.80 & 75.50 $\pm$ 0.80\\
MetaOptNet-SVM~\cite{mlwd}  	& ResNet-12 & Meta  & 62.64 $\pm$ 0.61 & 78.63 $\pm$ 0.46\\

%\noalign{\smallskip}
\midrule
%\noalign{\smallskip}
Matching Net~\cite{match-net}    & Conv-64 & Metric  & 43.56 $\pm$ 0.84 & 55.31 $\pm$ 0.73\\
ProtoNet~\cite{protonet} & Conv-64 & Metric  & 49.42 $\pm$ 0.78 & 68.20 $\pm$ 0.66\\
RelationNet~\cite{relationnet}  & Conv-64 & Metric   & 50.44 $\pm$ 0.82 & 65.32 $\pm$ 0.70\\
ProtoNet~\cite{protonet} & ResNet-12 & Metric  & 56.52 $\pm$ 0.45 & 74.28 $\pm$ 0.20\\
TADAM~\cite{tadam}  & ResNet-12 & Metric   & 58.50 $\pm$ 0.30 & 76.70 $\pm$ 0.38\\
Baseline++~\cite{closer-look}   & ResNet-18 & Metric   & 51.87 $\pm$ 0.77 & 75.68 $\pm$ 0.63\\
SimpleShot~\cite{simpleshot} & ResNet-18 & Metric & 62.85 $\pm$ 0.20 & 80.02 $\pm$ 0.14 \\
CMT~\cite{task-relevant}   & ResNet-18 & Metric   & 64.12 $\pm$ 0.82 & 80.51 $\pm$ 0.13\\
% AM3(ProtoNet, BERT\textsubscript{mask}) & ResNet-12 & Metric   & 62.11 $\pm$ 0.39 & 74.72 $\pm$ 0.64\\
AM3(ProtoNet, {GloVe}) & ResNet-12 & Metric   & 62.43 $\pm$ 0.80 & 74.87 $\pm$ 0.65\\
AM3(ProtoNet++)~\cite{AM3}  & ResNet-12 & Metric & 65.21 $\pm$ 0.49 & 75.20 $\pm$ 0.36\\
TRAML(ProtoNet)~\cite{traml}  & ResNet-12 & Metric & 60.31 $\pm$ 0.48 & 77.94 $\pm$ 0.57\\
CAN~\cite{CAN}  & ResNet-12 & Metric & 63.85 $\pm$  0.48 & 79.44  $\pm$ 0.34\\
DSN-MR~\cite{DSN-MR}  & ResNet-12 & Metric & 64.60 $\pm$ 0.48 & 79.51 $\pm$ 0.50\\
FEAT~\cite{feat}       & ResNet-12 & Metric & 66.78 & 82.05 \\
DeepEMD~\cite{deepemd}  & ResNet-12 & Metric & 65.91 $\pm$ 0.82 & 82.41 $\pm$ 0.56\\

%\noalign{\smallskip}
\midrule
%\noalign{\smallskip}
Ours(ProtoNet)      &  ResNet-12 &  Metric &  64.53 $\pm$ 0.37   &  79.37 $\pm$ 0.64\\
Ours(AM3,ProtoNet)  &  ResNet-12 &  Metric &  {\bf 68.42 $\pm$ 0.51}   &  81.29 $\pm$ 0.59\\
Ours(FEAT)       & ResNet-12 & Metric   &     67.84 $\pm$ 0.45 &  83.17 $\pm$ 0.72\\
Ours(DeepEMD)    & ResNet-12 & Metric   &  67.03 $\pm$ 0.79      &  {\bf 83.68 $\pm$ 0.65}\\
									
\bottomrule
\noalign{\smallskip}
\end{tabular}

\label{table4}
\end{center}
\end{table*}

\subsection{Experimental results}\label{secComparison}
AM3~\cite{AM3} and TRAML~\cite{traml} are the most direct competitors of our method, as these models also use class name embeddings. For this reason, we first present a detailed comparison with these methods in Section \ref{secAM4TRAML}. Subsequently, in Section \ref{secComparisonSOTA} we present a more general comparison with the state-of-the-art in few-shot learning.

\subsubsection{Comparison with AM3 and TRAML}\label{secAM4TRAML}
The comparison with AM3 can be found in Table \ref{table8a}, where we also show the impact of different types of class name embeddings. As can be seen, our proposed method outperforms AM3 in all cases, both in the 1-shot and 5-shot setting. This confirms the usefulness of decoupling the visual and textual prototypes, as this is the key difference between our model and AM3 when low-dimensional vectors, such as those from the GloVe model, are used. Furthermore, we can see that AM3 is not able to take advantage of the higher-dimensional embeddings, with the results for BERT, CON$_1$ and CON$_2$ all being worse than those for GloVe. This can be explained from the observation that these higher-dimensional class name embeddings result in a substantially higher number of parameters in the case of AM3, leading to overfitting. In contrast, thanks to the correlation exploration module, our method can exploit the additional semantic information that is encoded in the higher-dimensional embeddings without introducing any additional parameters in the classification model. In both the 1-shot and 5-shot settings, our model achieves the best results with CON$_2$ embeddings, which is in accordance with our findings from Section \ref{secAblationStudy}.

\begin{table}[t]
%\footnotesize
\begin{center}
\caption{The mean accuracies~(\%) with a 95\% confidence interval %of the 5-way 5-shot and 5-way 1-shot tasks 
on the CUB dataset.}
\begin{tabular}{llcc}
\toprule
%\noalign{\smallskip}
\textbf{Method} & \textbf{Backbone}
& \textbf{5-way 1-shot} & \textbf{5-way 5-shot}\\
%\noalign{\smallskip}
\midrule
%\noalign{\smallskip}
MAML          & Conv-64 &  55.92 $\pm$ 0.95 & 72.09 $\pm$ 0.76\\
Matching Net  & Conv-64 &  61.16 $\pm$ 0.89 & 72.86 $\pm$ 0.70\\
ProtoNet      & Conv-64 &  51.31 $\pm$ 0.91 & 70.77 $\pm$ 0.69\\
RelationNet   & Conv-64 &  62.45 $\pm$ 0.98 & 76.11 $\pm$ 0.69\\
Baseline++    & Conv-64 &  60.53 $\pm$ 0.83 & 79.34 $\pm$ 0.61\\
SAML~\cite{saml}   & Conv-64 &   69.35 $\pm$ 0.22 & 81.37 $\pm$ 0.15\\
DN4~\cite{revisit} & Conv-64 &  53.15 $\pm$ 0.84 & 81.90 $\pm$ 0.60\\
AM3(ProtoNet) & Conv-64 &  57.26 $\pm$ 0.66 & 71.34 $\pm$ 0.93\\
AM3(ProtoNet)~\cite{AM3}  & ResNet-12   &  73.6   &  79.9 \\
%\noalign{\smallskip}
\midrule
Ours(ProtoNet)      & Conv-64   &  69.79 $\pm$ 0.73  & 83.06 $\pm$ 0.66   \\
Ours(AM3,ProtoNet)  & Conv-64   &  72.14 $\pm$ 0.68  & 83.14 $\pm$ 0.69 \\
Ours(ProtoNet)     &  ResNet-12 &  76.58 $\pm$ 0.82  & 87.11 $\pm$ 0.71 \\
Ours(AM3,ProtoNet) &  ResNet-12 & {\bf 77.03 $\pm$ 0.85}  & {\bf 87.20 $\pm$ 0.70} \\
\bottomrule
\noalign{\smallskip}
\end{tabular}

\label{table5}
\end{center}
\end{table}

Regarding the TRAML model, as we did not have access to the source code, we only compare our method against the published results from the original paper \cite{traml}. As the base method, they considered both ProtoNet and AM3. As can be seen in Table \ref{table8b}, our method outperforms TRAML in both of these settings, for 1-shot as well as 5-shot learning. This is even the case if GloVe vectors are used for our model, although the best results are obtained when using the CON$_2$ embeddings for our model, while still using the GloVe vectors for the AM3 base model.

\subsubsection{Comparison with the State-of-the-Art}\label{secComparisonSOTA}
Tables \ref{table4}, \ref{table5} and \ref{table6} compare our model with existing methods on the miniImageNet, CUB and tieredImageNet datasets respectively, where miniImageNet and tieredImageNet are standard benchmarks for few-shot learning. CUB, which consists of 200 bird classes, allows us to evaluate the performance of our model on finer-grained classes. The performance of all methods is generally impacted by the choice of the backbone network. To allow for a fair comparison with different published results from the literature, in the case of miniImageNet, 
we show results of our model with ResNet-12 as the backbone, where possible (i.e.\ unless no published results are available for ResNet-12).
The results of the baselines in Table~\ref{table4} (miniImageNet) are obtained from \cite{traml}, \cite{feat}, \cite{DSN-MR}, \cite{CAN} and \cite{deepemd}. The results for the baselines in Table~\ref{table5} (CUB) are obtained from \cite{saml},  \cite{revisit} and \cite{AM3}. These results are based on the Conv-64 and ResNet-12 backbone, which we therefore adopt as well for this dataset. The results for tieredImageNet in Table \ref{table6} primarily rely on ResNet-12 as backbone, where the baseline results have been obtained from \cite{feat}, \cite{AM3}, \cite{CAN} and~\cite{DBLP:conf/eccv/TianWKTI20}. 
Apart from changes to the backbone network, we also vary the base method that is used as the visual classification component of our model. We have used ProtoNet, AM3 (with ProtoNet and GloVe vectors), FEAT and DeepEMD for this purpose.

The results in Table \ref{table4} show that when ProtoNet is used as the base model, our method substantially outperforms the standard ProtoNet model, with the accuracy increasing from 56.52 to 64.53 in the 1-shot setting and from 74.28 to 79.37 in the 5-shot setting. Similarly, when using AM3, FEAT and DeepEMD as the base model, the results improve on the standard AM3, FEAT and DeepEMD models, respectively. The versions of our model with AM3 and DeepEMD also achieve the best overall results for the 1-shot and 5-shot settings respectively.
The results for CUB in Table \ref{table5} again show that our model is able to substantially outperform the standard ProtoNet model. We also find that our model outperforms AM3, with the best results obtained when combining our model with AM3. In addition to the Conv-64 backbone, we have also included results with ResNet-12 for our model and AM3, which confirm these conclusions.
Finally, for the tieredImageNet results in Table \ref{table6}, we again see that our method consistently leads to improvements of the base model. In particular, this is shown for four different choices of the base model: ProtoNet, AM3, FEAT and DeepDEM. The version of our model that is based on DeepEMD leads to the best results overall.

\section{Conclusions}
We have proposed a method to improve the performance of metric-based FSL approaches by taking class names into account. Experiments on three datasets show that our method consistently improves the results of existing metric-based models. Moreover, our method is conceptually simple and can easily be added to a wide range of (existing and future) FSL models. An important advantage compared to previous work on exploiting class name embeddings, such as the AM3 method, is that we do not have to increase the number of parameters of the classification model. This has allowed us to exploit higher-dimensional class name embeddings. In particular, we have used class name embeddings that were learned using the BERT masked language model, as well as concatenations that combine different types of embeddings. From a technical point of view, our approach relies on two key insights. First, we found that decoupling the visual and textual prototypes is essential to achieving good results. Second, to avoid the introduction of new parameters, we rely on variants of canonical correlation analysis to align class name embeddings with the corresponding visual prototypes.

\begin{table}[t]
%\footnotesize
\begin{center}
\caption{The mean accuracies~(\%) with a 95\% confidence interval %of the 5-way 5-shot and 5-way 1-shot tasks 
on the tieredImageNet dataset. %Note CMT and SimpleShot~\cite{simpleshot} utilize the ResNet-18, the rest utilize the ResNet-12
}
\begin{tabular}{llcc}
\toprule
%\noalign{\smallskip}
\textbf{Method} & \textbf{Backbone}
& \textbf{5-way 1-shot} & \textbf{5-way 5-shot}\\
%\noalign{\smallskip}
\midrule
%\noalign{\smallskip}
ProtoNet    & ResNet-12  & 53.31 $\pm$ 0.89 & 72.69 $\pm$ 0.74\\
RelationNet & ResNet-12  & 54.48 $\pm$ 0.93 & 71.32 $\pm$ 0.78\\
MetaOptNet  & ResNet-12  & 65.99 $\pm$ 0.72 & 81.56 $\pm$ 0.63\\
CTM         & ResNet-18  & 68.41 $\pm$ 0.39 & 84.28 $\pm$ 1.73\\
SimpleShot  & ResNet-18  & 69.09 $\pm$ 0.22 & 84.58 $\pm$ 0.16\\
AM3(ProtoNet)    & ResNet-12 & 58.53 $\pm$ 0.46 & 72.92 $\pm$ 0.68\\
AM3(ProtoNet++)  & ResNet-12 & 67.23 $\pm$ 0.34 & 78.95 $\pm$ 0.22\\
CAN         & ResNet-12  & 69.89 $\pm$ 0.51 & 84.23 $\pm$ 0.37\\
FEAT        & ResNet-12  & 70.80 $\pm$ 0.23 & 84.79 $\pm$ 0.16\\
DeepEMD     & ResNet-12  & 71.16 $\pm$ 0.87 & 86.03 $\pm$ 0.58\\
Rethinking~\cite{DBLP:conf/eccv/TianWKTI20}  & ResNet-12  & 71.52 $\pm$ 0.69 & 86.03 $\pm$ 0.49 \\
%\noalign{\smallskip}
\midrule
%\noalign{\smallskip}
Ours(ProtoNet) & ResNet-12 & 66.82 $\pm$ 0.65         &    78.97 $\pm$ 0.53 \\
Ours(AM3,ProtoNet) &  ResNet-12   & 67.22 $\pm$ 0.43  &    79.08 $\pm$ 0.58 \\
Ours(FEAT)     & ResNet-12 & 72.31 $\pm$ 0.68         &    85.76 $\pm$ 0.36 \\
Ours(DeepEMD)  & ResNet-12 & {\bf 73.76 $\pm$  0.72}  &    {\bf 87.51 $\pm$ 0.75}\\
\bottomrule
\noalign{\smallskip}
\end{tabular}

\label{table6}
\end{center}
\end{table}

\begin{acks}
	This research was supported in part by the National Key R\&D Program of China (2017YFB1200700); Capital Health Development Scientific Research Project (Grant 2020-1-4093); Clinical Medicine Plus X - Young Scholars Project, Peking University, the Fundamental Research Funds for the Central Universities; Global Challenges Research Fund (GCRF) grant (Essex reference number: GCRF G004); HPC resources from GENCI-IDRIS (Grant 2021-[AD011012273] and ANR CHAIRE IA BE4musIA.
\end{acks}

\bibliographystyle{ACM-Reference-Format}
\bibliography{sample-base}

\appendix
\section{Appendix} 

We now explain in more detail how the matrices $\mathbf{A}$ and $\mathbf{B}$ are constructed. Let $\mathbf{X_0}$ and $\mathbf{Y_0}$ be the matrices whose i\textsuperscript{th} row is, respectively, the class name embedding and the visual prototype of the i\textsuperscript{th} class. The visual prototypes in $\mathbf{Y_0}$ are estimated by averaging the visual features $f_{\theta}(x)$ of all images $x$ from the training set that belong to the i\textsuperscript{th} class. These visual prototypes thus differ from those that are used for training the main model, as they are estimated from the full training set, rather than from a sampled episode.

As pointed out by \cite{DBLP:conf/aaai/ArtetxeLA18}, we can think of alignment methods such as CCA as performing a sequence of linear transformation steps. In particular, to find the matrices $\mathbf{A}$ and $\mathbf{B}$, we can use the following steps. The first transformation, called whitening, ensures that the individual components of the vectors have unit variance and are uncorrelated:
\begin{align*}
\mathbf{X_1} &= \mathbf{X_0}\mathbf{A_1}&
\mathbf{Y_1} &= \mathbf{Y_0}\mathbf{B_1}
\end{align*}
where
\begin{align*}
\mathbf{A_1} &= (\mathbf{X_0}^T\mathbf{X_0})^{\frac{1}{2}} &
\mathbf{B_1} &=(\mathbf{Y_0}^T\mathbf{Y_0})^{\frac{1}{2}}
\end{align*}
The second transformation maps the two embedding spaces onto a shared space using two orthogonal transformations $\mathbf{A_2}$ and $\mathbf{B_2}$. In particular, let us write  the singular value decomposition of $\mathbf{X_1}^T\mathbf{Y_1}$ as $\mathbf{A_2}\mathbf{S}\mathbf{B_2^T}$. Then we have  $\mathbf{X_2} = \mathbf{X_1}\mathbf{A_2}$ and $\mathbf{Y_2} = \mathbf{Y_1}\mathbf{B_2}$. If de-whitening is used, the next transformation aims to restore the initial variances and correlations, i.e.\ we have $\mathbf{X_3} = \mathbf{X_2}\mathbf{A_3}$ and $\mathbf{Y_3} = \mathbf{Y_2}\mathbf{B_3}$, where:
\begin{align*}
\mathbf{A_3}  = \mathbf{A_2}^{T}\mathbf{A_1}^{-1}\mathbf{A_2}\\
\mathbf{B_3}  = \mathbf{B_2}^{T}\mathbf{B_1}^{-1}\mathbf{B_2}
\end{align*}
The final step is dimensionality reduction. Let $\mathbf{A_4}$ be the $m_t \times d$ matrix whose i\textsuperscript{th} row has a 1 in the i\textsuperscript{th} column and 0s everywhere else, and similar for the $m_v \times d$ matrix $\mathbf{B_4}$.

In summary, the transformation $\mathbf{A}$ of the class name embedding space is given by $\mathbf{A} = \mathbf{A_1}\mathbf{A_2}\mathbf{A_3}\mathbf{A_4}=\mathbf{A_2}\mathbf{A_4}$ if de-whitening is used and by $\mathbf{A} = \mathbf{A_1}\mathbf{A_2}\mathbf{A_4}$ if standard CCA is used. Similarly, the transformation $\mathbf{B}$ of visual prototype space is given by $\mathbf{B} = \mathbf{B_1}\mathbf{B_2}\mathbf{B_3}\mathbf{B_4}=\mathbf{B_2}\mathbf{B_4}$ if de-whitening is used and by $\mathbf{B} = \mathbf{B_1}\mathbf{B_2}\mathbf{B_4}$ otherwise.

\end{document}